# Investigating the Performance of Correspondence Algorithms in Vision based Driver-Assistance in Indoor Environment

F.Mahmood
College of E&ME
(NUST) Peshawar Road
Rawalpindi

Syed.M.B.Haider
College of E&ME
(NUST) Peshawar Road
Rawalpindi

F.Kuwar
College of E&ME
(NUST) Peshawar Road
Rawalpindi

## ABSTRACT
This paper presents the experimental comparison of fourteen stereo matching algorithms in variant illumination conditions. Different adaptations of global and local stereo matching techniques are chosen for evaluation The variant strength and weakness of the chosen correspondence algorithms are explored by employing the methodology of the prediction error strategy. The algorithms are gauged on the basis of their performance on real world data set taken in various indoor lighting conditions and at different times of the day.

## Keywords
Performance, Indoor, Lighting, Correspondence, Algorithms

## 1. INTRODUCTION
In mobile robotics and unmanned vehicles the technique of stereo vision in mainly applied to find the distances and 3D information. The calculation of an accurate depth or disparity values is vital because it is the basic step of understanding the surrounding environment of an autonomous vehicle e.g. obstacle avoidance and navigation. A given correspondence algorithm may report confusing depth values if the frames are not acquired under ideal lighting environments. An algorithm is required which proves efficient in all the challenging illumination conditions and will benefit many applications including robotics, rescue and rehabilitation.

A large number of algorithms are proposed and most of the papers do not quantitatively compare their own presented method with the previous approaches. The performance evaluation of correspondence algorithms is widely discussed in [2, 3, 8-10]. The common approach employed in the above mentioned papers is that they use the images of Middlebury database [1] in their assessment. The greatest drawback in their evaluation is that the Middlebury images are taken under controlled lighting conditions. Hence the gained result in these evaluations is only of limited use as there is significant lighting variation in the real world. The papers in [4, 5, 11 and 12] evaluated the efficiency of algorithms facing the challenges of outdoor world environment. To our knowledge very little evaluation has been yet performed on how the variation of lighting conditions influence the performance of correspondence algorithms in indoor environment, which is the major contribution of this paper.

This report throws light on the influence of three different indoor lighting conditions (fluorescent lighting condition, incandescent lighting condition and day light environment) on fourteen correspondence algorithms. The main aim of this task was to find the best performing correspondence algorithm for our vision based IDRIS (Intelligent DRIving System) facing the indoor environment. The set of chosen algorithms are adaptations and different parameterizations of global and local stereo matching techniques which will be explained in subsequent subsection. Other correspondence algorithms could also be considered as well, but this set provides a good selection of currently preferential stereo matching approaches. This report likes to point out that each matching technique behaves uniquely in alternative indoor illumination environments.

This paper is structured as follows. Problem Statement is discussed in section 2. Methodology is explained in section 3. Brief description of the selected algorithms is presented in section 4. Experimental evaluation is discussed in section 5. Section 6 shows the conclusion. References are shown at the end.

## 2. PROBLEM STATEMENT
In case of matching problem it is well known that the key requirement for the success of the correspondence algorithm is the level of contrast in the images to be matched. Homogeneous or texture less scenes give bad results for most algorithms under any lighting conditions. Basically the stereo system's aperture and exposure time is adjusted to ensure every level of contrast in the images obtained. The most discernable physical difference between used lighting conditions is the fluctuations present in three lights: fluorescent light (50-60) Hz, tungsten filament (20-30) Hz whereas the day light is constant. Relative intensities of the 3 light sources are also a big problem if the used lenses are not capability of optical light adjustment. The images were also included in which the used optical system was not perfectly synchronized. The matching problem gets tough in the presence of low lighting conditions i.e. in the dark areas after sunset. Besides finding an efficient algorithm this paper also gives information about choosing the algorithm in the limited lighting environment as it is hard to decide which algorithms should be used for different applications or to decide which algorithm performs the best under certain situations [7].

## 3. METHODOLOGY
To gauge the performance of correspondence algorithms facing real world challenges, a 3 stage analysis is used. Fig. 1 shows the sequence of steps for evaluating the algorithms. In the first stage, images are acquired from our correspondence rig and disparity maps are computed. In the second stage, the prediction error strategy is applied and finally the quantitative evaluation is presented in the last stage. An allowable tolerance is given to all the algorithms during their testing. The used metric and the test data are explained as below.

## 3.1 Collection of Datasets

In the first stage of our analysis images are acquired from a locally developed correspondence rig. Images are of resolution (800x600) pixels. For a stereo vision system, performance depends on many factors, including the lightening conditions, noise level in the images, the true range of objects in the scene and the illumination artifacts introduced by matching algorithms. Also, shadows affect the performance of a correspondence algorithm significantly, this effect can be minimized by increasing the number of light sources illuminating the scene, but too much light also affects the performance of a correspondence algorithm adversely. Sunlight from windows also affects the performance of correspondence algorithms. All these factors were kept in mind while acquiring our dataset. The dataset is divided in 3 cases, each having its significance in terms of lightening and range of objects in the scene.

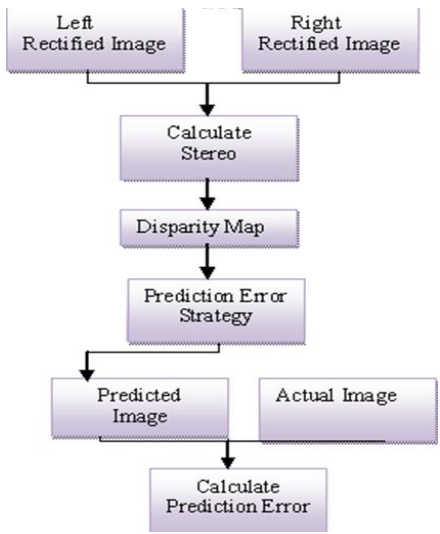

**Figure 1 Flow chart showing different stages of our analysis**

- Case 1: Fluorescent light from different indoor illumination sources in the absence of sunlight that is, during night time.

- Case 2: Day light environment in the presence of sunlight. These images were taken at different times of the day that is, morning, noon and evening to enhance the scope of our analysis.

- Case 3: Incandescent light from different indoor illumination sources in the absence of sunlight that is, during night time.

The purpose of taking such a challengeable dataset is, to not leave any loopholes in our analysis. Fig. 2 shows sample images used in our experimentation.

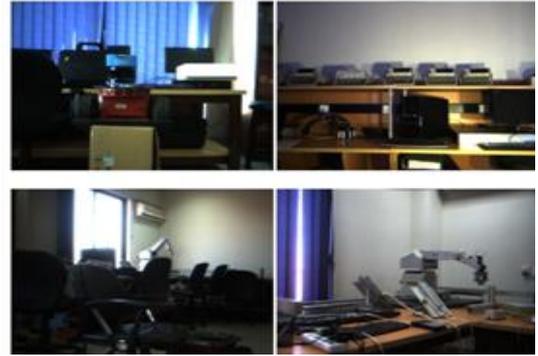

**Figure 2: Sample images used in our datasets**

## 3.2 Prediction Error Strategy

Our methodology uses the concept of prediction error strategy for measuring the quality of correspondence algorithms. It works in equally efficient way where the ground truth is not available or quite difficult to acquire with full accuracy. The presented methodology evaluates the capability of disparity map to foresee an (unseen) third image. The third image is taken from the identified camera position with respect to the input pair of images. To apply this methodology to stereo matching, first the depth map is computed using the given input pair of images then measures how accurately the depth map and the input pair of images predicts the third image. The prediction error will be evaluated by comparing the synthesized view against the actual view as inferred from [22]. See Fig.3 for an example of recorded third view, synthesized view and the disparity map. The disparity map was obtained by applying ELAS correspondence algorithm.

Inverse warping was employed for the prediction of the third image as no decision has to be made as to which the gaps need to be filled. It pulls pixels from an unseen view back into the coordinate frame of the original reference image. SAD algorithm was finally employed as a quality metric for computing the degree of uniformity between the $3^{rd}$ image and the synthesized view as mentioned in Eq.1.

$$E(t) = \sum_{i,j \in \cup} |I_1(x+i, y+j) - I_2(x+dx+i, y+dy+j)| \quad (1)$$

In Eq.1 $I_1$ corresponds to the warped image and $I_2$ corresponds to the third image. E(t) is the error image formed from the sum of absolute difference (SAD) equation.

$$m_N = \frac{1}{T} \sum_{t=1}^{T} N(t) \quad (2)$$

In Eq.2 N(t) is the percentage of the accurately matched pixels in each frame. T is the total number of images used. The mean $m_N$ will allow us to identify the best performing algorithm in each case.

$$\sigma^2 = \frac{1}{T} \sum_{t=1}^{T} (N(t) - m_N)^2 \quad (3)$$

In Eq.3 $\sigma^2$ shows the variance which shows the steadiness of each algorithm. T is the number of total images used in each case.

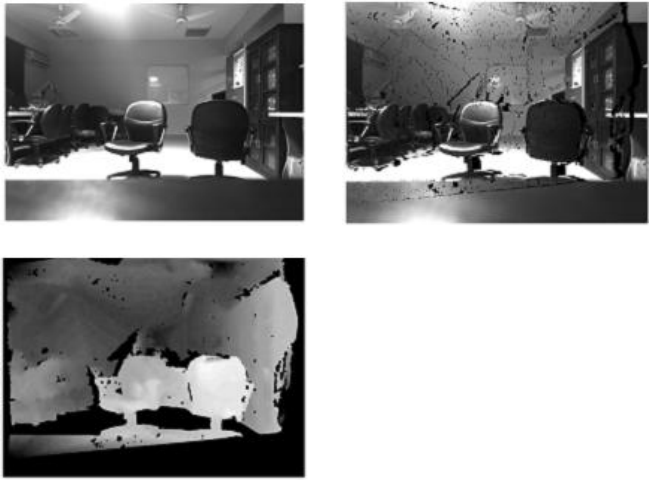

**Figure 3 (Up-LEFT) Actual Image view. (UP-RIGHT) Predicted image for the disparity shown on the Bottom. The matching algorithm applied was LIBEAS**

## 4. CORRESPONDENCE ALGORITHMS

### 4.1 Global Energy Minimization (GEM)
GEM [19] belongs to a class of region based correspondence matching technique which uses block matching technique to construct an error energy matrix for every disparity. Averaging filter is applied iteratively to every energy matrix which removes very sharp change in energy which could belong to incorrect matching. At last the minimum error energy is found and the disparity index is assigned to every pixel which will build an accurate and reliable disparity map.

### 4.2 Line Growing Based Stereo Matching (LGBS)
This algorithm [19] is based on region growing process in which the root point is found if its energy is less or equal to the specified threshold. In the region growing process the region grows from the stereo correspondence point to a predefined value in the direction of rows only. In the growing process it calculates the error energy of the neighboring pixels and associate that point to the region if its error energy is less than the predetermined threshold energy. This process is done repeatedly until all the pixels of the image are processed. Finally the grown disparity regions compose of disparity map.

### 4.3 Efficient Large Scale stereo Matching (ELAS)
This algorithm [18] is based on a generative probabilistic model for correspondence matching. Firstly the disparity of a set of robustly matched correspondences named support points is calculated. The image coordinates of the support points are then used to create a 2D mesh via Delaunay triangulation. A prior is computed to disambiguate the matching problem and is formed by computing a piecewise linear function induced by the support point disparities and the triangulated mesh.

### 4.4 Segment-Based Stereo Matching (SFA)
SFA uses the approach of segment based correspondence matching technique [20] in which regions of homogeneous color are located by applying a color segmentation method. Self-adapting dissimilarity measure is then used to increase the number of reliable correspondences. An insensitive approach is applied to extract the disparity planes. Instead of assigning a disparity value to each pixel, a disparity plane is assigned to each segment. Finally the optimum disparity labeling problem is solved using belief propagation algorithm.

### 4.5 Belief Propagation Image Matching (BP)
This is an update version of belief propagation algorithm [16] in which new techniques are applied that substantially improve the running time of belief propagation (BP) for solving early vision problems. The complexity of this algorithm is reduced to be linear rather than quadratic in the number of possible labels. One technique makes it possible to obtain the good results with a small fixed number of message passing iterations independent of the size of the input image. Taken together these techniques speed up this algorithm making its running time competitive with local methods.

### 4.6 Growing Corresponding Seeds algorithm (GCS)
GCS [13] forms a semi dense disparity map by visiting only a small fraction of disparity space. It grows from small set of few random seed correspondence correspondences in which each seed is considered as a point in disparity space. Then global optimization technique is then applied to find optimal matching with the initially found correspondence correspondences, thereby making the algorithm more fast and robust. This algorithm has the capability to recover if the initial correspondence correspondences are wrong and proves efficient even in wide base line correspondence which guarantees matching accuracy and correctness even in the repetitive patterns at a small computational cost.

### 4.7 Fast Bilateral Stereo Matching (FBS)
FBS [12] is a local correspondence algorithm in which a new cost aggregation strategy is proposed based on joint bilateral filtering and incremental calculation schemes. In bilateral filtering both geometric (spatial filter) and a color proximity constraint (range filter) is jointly and independently enforced. This strategy reduces the computation cost, allowing the computation of an efficient and accurate disparity map.

### 4.8 Graph Cut Algorithm (GC)
This algorithm [17] presents two methods (expansion move and swap move) which properly address occlusions, while preserving the advantages of graph cut algorithms. In this algorithm a disparity- $\alpha$ (in a fixed order or at random) is selected and a unique configuration is found within a single $\alpha$-expansion move If this decreases the energy, then it goes there; it stops if there is no $\alpha$ that decreases the energy. The critical step in this method is to efficiently compute the $\alpha$-expansion with the smallest energy and graph cuts helps to solve this problem.

## 4.9 Local Correspondence matching techniques

These matching processes involve the computation of similarity measure for each disparity value followed by an aggregation and optimization steps. Aggregation step is achieved by taking a square window of certain size around the pixel of interest in the reference image and finding the homologous pixel within the window in the target image. Winner take all strategy is applied as the optimization step. Following are four different similarity measures [8] having the same aggregation and optimization steps which are as follows.

### 4.9.1 Sum of absolute difference (SAD)

$$SAD = \sum_{(i,j \in \cup)} \left| I_1(x+i, y+j) - I_2(x+dx+i, y+dy+j) \right| \quad (4)$$

Sum of Absolute Differences (SAD) is one of the simplest of similarity measures which is calculated by subtracting pixels within a square neighborhood between the reference image $I_1$ and $I_2$. Aggregation is done by taking the absolute difference within the square window.

### 4.9.2 Sum of squared difference (SSD)

$$SSD = \sum_{(i,j \in \cup)} \left( I_1(x+i, y+j) - I_2(x+dx+i, y+dy+j) \right)^2 \quad (5)$$

In (SSD) Sum of squared difference the difference between the pixels are squared and aggregated within the squared window.

### 4.9.3 Sum of hamming difference (SHD)

$$SHD = \sum \left( I_1(x, y) \, XOR \, I_2(x+i, y+j) \right) \quad (6)$$

SHD is normally employed for matching census-transformed images by computing bitwise-XOR of the values in the left and right images $I_1$ and $I_2$ respectively.

### 4.9.4 Normalized Cross correlation (NCC)

$$NCC = \frac{\sum_{(i,j \in \cup)} I_1(i,j) \bullet I_2(x+i, y+j)}{\sqrt{\sum_{(i,j) \in \cup} I_1(i,j)^2 \bullet \sum_{(i,j) \in \cup} I_2(x+i, y+j)}} \quad (7)$$

NCC involves numerous multiplication, division and square root operations applied on the pixels of image $I_1$ and $I_2$.

## 4.10 Enhanced Normalized Cross correlation (ENCC)

This algorithm [14] is a correlation based similarity measure which is based on a specific linear interpolation scheme on the intensities of two adjacent candidate windows. It is not only capable of producing the disparity map with sub pixel accuracy but also the feature of invariance in photometric distortions is incorporated in it.

## 4.11 Connectivity Slant Algorithm (MSA)

This algorithm [15] examines the implication of shape on the process of finding dense stereo correspondence and half-occlusions for correspondence images. The desired property of the disparity map is that it should be a piecewise continuous function with minimum number of discontinuities. It introduces the horizontal slanted surface (i.e., having depth variation in the direction of the separation of the two cameras) and vertical slanted surface to create a first order approximation to piecewise continuity.

## 5. EVALUATION FOR EACH CASE

This section will illustrate the use of cases to evaluate the performance of correspondence algorithms for each case. Table 1 will demonstrate the performance of every algorithm in our developed cases. Fig 4 shows complete diagrams of performance of correspondence algorithms in each case. In this diagram one can see that performance of algorithms may differ within one case from one frame to another.

## 5.1 Performance of correspondence algorithms for Case 1(fluorescent lighting condition)

This sequence is recorded after the sunset with different range of obstacles. 100 images are taken in this condition with various challenges including reflections, shadows, variant illumination with both near and far obstacles. For this particular sequence it was noticed that the near obstacles degrade the performance of the tested correspondence algorithms, causing more reflections in the imagery. The top performing algorithm was (GC) followed by ELAS and GEM. ELAS is based upon generative probabilistic model (Bayesian Approach) for stereo matching and follows an approach which builds a prior over the disparity space by forming a triangulation on a set of robustly matched correspondences. This results in an efficient algorithm that is robust against moderate changes in illumination and well suited for robotics application. . Fig 4a and 4b shows the complete overview of the tested correspondence algorithms in the fluorescent lighting condition.

## 5.2 Performance of correspondence algorithms for Case 2(day lighting condition)

This case covers the ordinary driving condition at the day time from morning till evening. 60 images were acquired both containing the sunny and cloudy day environment. In this condition the brightness differences between the stereo pair is low, the sun is still high in the sky, shadows and specularities are minimum, windows are open with inner lights switched off and the whole gallery is sun bleached. Most of the algorithms showed better performance in sunny day as compared to cloudy day environment because of having more shadows in the imagery. The algorithm with the better performance was GC followed by ELAS and GEM. The performance of all the algorithms degrades when the obstacles comes closer to the camera. Fig 4c and 4d depict the performance of the correspondence algorithms in day lighting condition.

## 5.3 Performance of correspondence algorithms for Case 3(Incandescent lighting condition)

The images in this condition were taken both at night time and during sunset. The indoor environment was lightened with incandescent light, reflections were less and the brightness difference between the stereo pair was present. Brightness Difference is a common issue in autonomous driving vehicles while changing the visual angle or due the non-uniform distribution of light intensity in the surrounding environment. One interesting point to note with this case is that BP had the best performance however the performance of this algorithm was less in other two cases. The better performance of BP in the presence of brightness difference is due to intensity consistency in the data term of its cost function. This ranking is similar to the results obtained in [22] in the case for the brightness difference sequence in which the results of BP was improved, supporting the idea that the prediction error is a good technique to assess stereo algorithms in the absence of ground truth. GC algorithm ranked fifth. The worst performance was showed by the GCS. Fig 4e and 4f describes the performance of the tested correspondence algorithms in incandescent lighting condition.

## 5.4 Which is the best algorithm in indoor environment?

Using the mean of SAD over all lighting environments Graph-Cut outperforms all other algorithms with the mean accuracy of 92.6. It performs the best in two situations and lagged in the third case (incandescent lighting condition) by a small difference. The main advantage of GC is having the strong smoothness penalties which come from its construction. The max-flow/min-cut algorithm allows the Graph Cuts algorithm to compute the optimal swap for the whole graph. This global computation helps GC to handle with the strong interactions and to surpass its performance than the other algorithms.

## 6. CONCLUSION

The best suitable algorithm for our manually developed correspondence rig in Intelligent Driving System (IDRIS) was found. For this we collected the dataset of variant lighting condition including day light (morning, noon and evening), fluorescent light and incandescent light. We tested 14 correspondence algorithms on our evaluation process. Prediction Error Strategy is applied in which the virtual image is created with the help of input images and the computed disparity map. Then the deviation of this image is calculated from the third image. Although Graph Cut algorithm proved best in only two cases even then it is the best performed algorithm in the overall evaluation. This paper only discusses three situations. Summarizing our more general observations and experiences we may conclude that the following cases hold.

- By extensive experimentation we have concluded that GC can handle the problems of occlusions and ordering constraint (if pixel / feature a is left to b the matching pixel / feature a' needs to be left to b') due to the presence of two energy moves (expansion move and swap move) in its algorithm. Moreover it has the capability to create convex regions in the situations having nearly constant depth.
- The main disadvantage of the cost function that they are sensitive to the brightness difference (because their dependence on brightness consistency).
- BP is a good choice if there is insignificant variation of depth in the imagery.
- The performance of SFA heavily depends on the amount/percentage of pixels showing a planar area

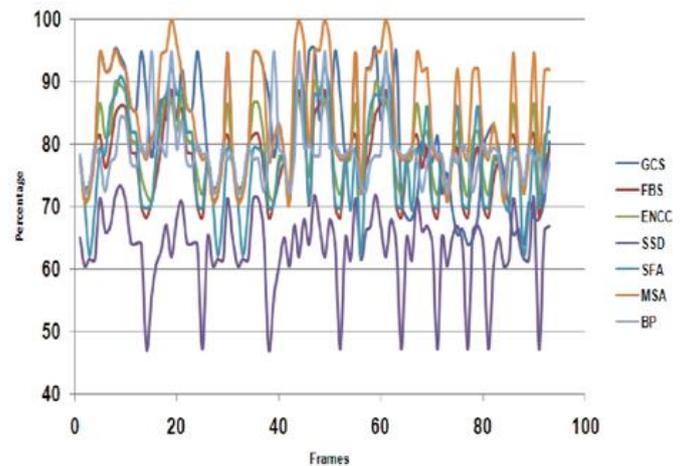

**Figure 4(a) Performance of 7 algorithms in florescent light**

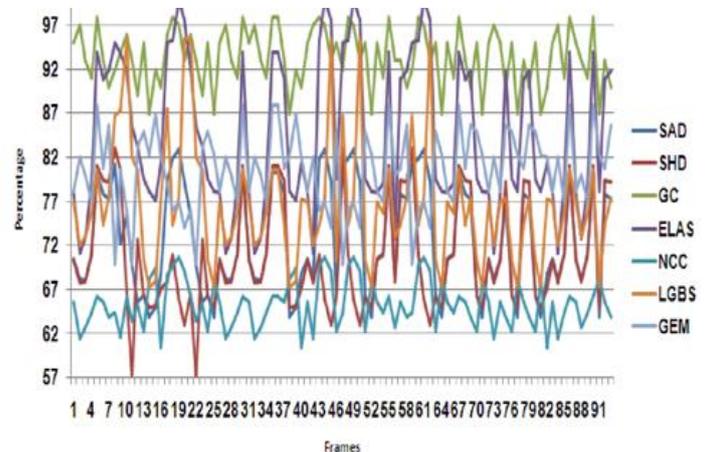

**Figure 4(b) Performance of 7 algorithms in florescent light**

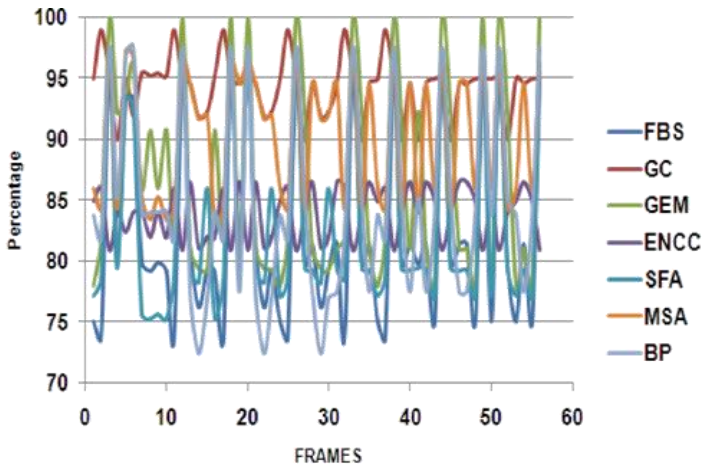

Figure 4(c) Performance of 7 algorithms in day light

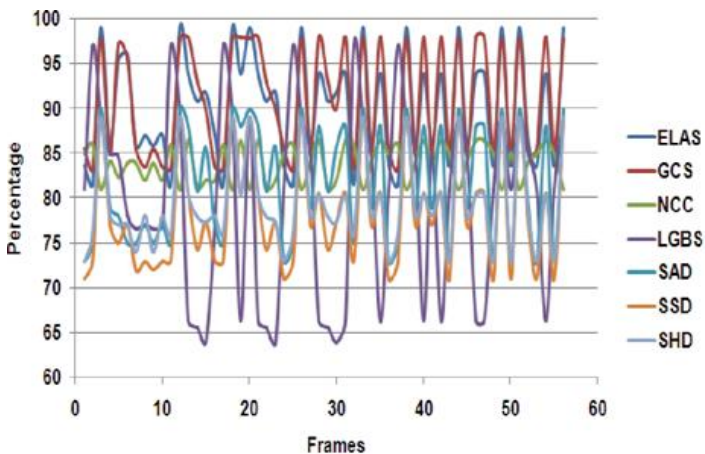

Figure 4(d) Performance of 7 algorithms in day light

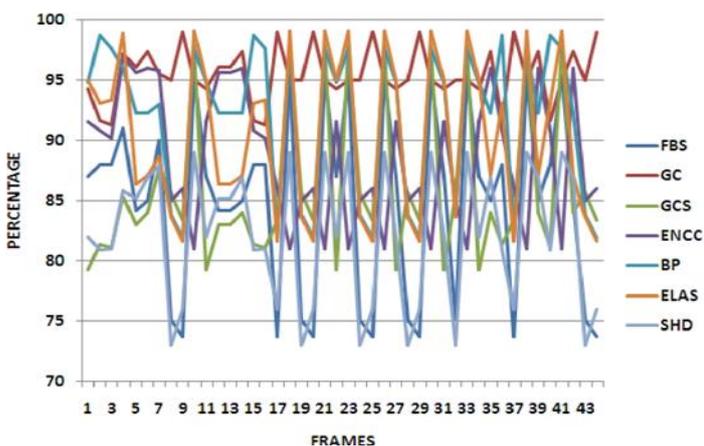

Figure 4(e) Performance of 7 algorithms incandescent light

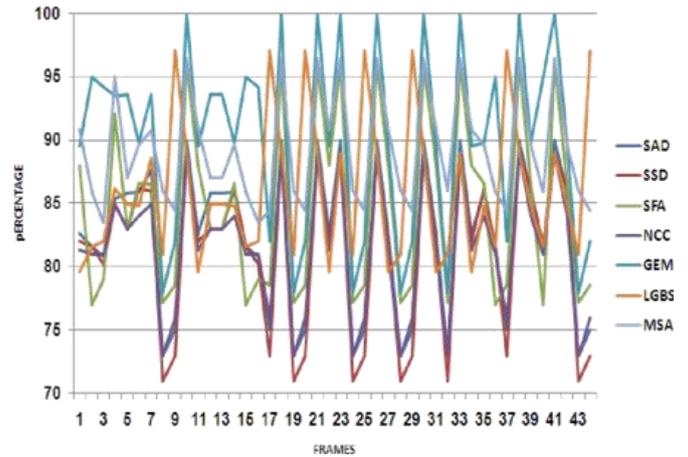

Figure 4(f) Performance of 7 algorithms in Incandescent light